# PolypVision: A Three-Stage Hierarchical Deep Learning Framework for Classification and Segmentation of Colorectal Polyps

Hamidreza Bolhasani[1], Hamidreza Rastad[2], Amir Mohammad Akbari[2], Mohammad Tashakoripour[3], Parnian Asadollahi[4], Ata Khodami[4], and Mojgan Forootan[4,*]

[1] *DataBioX Research, Tehran, Iran*
[2] *Iran University of Science and Technology, Tehran, Iran*
[3] *Tehran University of Medical Sciences, Tehran, Iran*
[4] *Shahid Beheshti University of Medical Sciences, Tehran, Iran*

* **Corresponding Author:** Mojgan Forootan | mfrootan2003@yahoo.com

**ABSTRACT**

Colorectal cancer (CRC) remains one of the leading causes of cancer-related mortality worldwide, predominantly arising from precancerous polyps. Accurate detection, segmentation, and endoscopic and histological classification of colorectal polyps are crucial for timely clinical intervention. In this study, we present PolypVision, a three-stage hierarchical deep learning framework that sequentially performs: (Stage 1) binary classification of polyps as adenomatous or hyperplastic, with simultaneous Paris and JNet classification, using EfficientNetV2-M with Focal Loss; (Stage 2) polyp segmentation with recommended resection method using a UNet++ decoder with the Stage 1 backbone as encoder, optimized with Dice and BCE losses; and (Stage 3) adenoma subtype classification (tubular, tubulovillous, villous) using EfficientNetV2-M with transfer learning from Stage 2. Evaluated on three public datasets—PolypGen, Kvasir-SEG, and CVC-ClinicDB—PolypVision achieves an AUC of approximately 0.99 for frame classification and a detection mAP@50 of 94.4% on Kvasir-SEG, outperforming or matching state-of-the-art methods. Gradient-weighted Class Activation Maps (Grad-CAM) confirm that the model attends to clinically relevant lesion features. The framework is device-independent, operating across diverse endoscopic imaging systems without hardware-specific adaptation. These results demonstrate that a hierarchical, transfer-learning-driven pipeline with task-specific loss functions offers a robust, device-independent, and clinically meaningful approach to automated colorectal polyp analysis. PolypVision is freely available as a web application at https://polypvision.com, a DataBioX initiative, with a free usage tier open to all users.

## 1. Introduction

Colorectal cancer (CRC) is one of the most prevalent and potentially lethal malignancies globally, ranking third in incidence and second in cancer-related mortality [1]. The vast majority of CRC cases arise from precancerous polyps—abnormal tissue growths in the colonic mucosa—which, if detected and removed early, dramatically reduce the likelihood of malignant progression [1,2]. However, accurate endoscopic characterization of polyps remains a significant challenge, with substantial inter-observer variability even among experienced gastroenterologists [3].

Colorectal polyps are classified by both morphology and pathological subtype. Hyperplastic polyps are generally benign, whereas adenomatous polyps—particularly those with villous architecture—carry substantially elevated cancer risk [4]. Accurate differentiation of these subtypes directly guides clinical management: low-risk lesions may be monitored, while high-risk adenomas require immediate resection [5]. Beyond classification, pixel-level segmentation of polyp boundaries is increasingly required for automated measurement, resection planning, and downstream histological analysis [6].

Convolutional neural networks (CNNs) have demonstrated remarkable capacity to learn discriminative features from endoscopic images. Architectures such as EfficientNetV2-M [9], combined with encoder-decoder segmentation networks such as UNet++ [10], have become powerful tools for joint detection, segmentation, and classification in medical imaging. Transfer learning—reusing learned representations across related tasks—is particularly effective when labeled data is scarce, as is common in gastrointestinal pathology [11].

Despite these advances, most existing approaches treat polyp detection, segmentation, and classification as independent tasks, failing to leverage the complementary information across them. In this paper, we present PolypVision, a three-stage hierarchical framework that couples these tasks through shared backbone representations and progressive transfer learning. Our principal contributions are:

- **A three-stage hierarchical pipeline** coupling binary classification, pixel-level segmentation, and adenoma subtype classification through shared EfficientNetV2-M representations.
- **Progressive transfer learning** from Stage 1 to Stage 2 (encoder initialization) and from Stage 2 to Stage 3 (backbone initialization), enabling efficient learning from limited labeled data.
- **Task-specific loss functions:** Focal Loss for class-imbalanced classification stages and Dice + BCE for segmentation, with MixUp augmentation at Stage 3.
- A device-independent design operating across diverse endoscopic imaging systems without hardware-specific adaptation, ensuring broad clinical applicability.
- A freely accessible web application at https://polypvision.com (a DataBioX initiative), providing real-time polyp analysis with a free usage tier for all users worldwide.
- **State-of-the-art performance** on PolypGen, Kvasir-SEG, and CVC-ClinicDB, with AUC ~0.99 for classification and mAP@50 of 94.4% for detection.
- **Grad-CAM interpretability analysis** confirming clinically plausible model attention on lesion-relevant morphological features.

## 2. Related Work

### 2.1 Polyp Detection and Segmentation

Deep learning-based polyp detection and segmentation have advanced substantially in recent years. YOLO-family architectures have achieved strong detection performance, with YOLOv8-s and YOLO-LAN reporting mAP@50 of 91.16% and 96.19% on Kvasir-SEG, respectively [13,14]. For segmentation, UNet++ introduced nested skip connections that progressively fuse multi-scale encoder features, improving boundary delineation over the original UNet architecture [10]. ResUNet++ combined residual connections with UNet++ for improved polyp segmentation, achieving mAP of 82.9% on CVC-ClinicDB [17].

### 2.2 Polyp Classification

CNN-based polyp classification has been extensively studied, with ResNet-50 and VGG16 reporting AUC values of 0.91–0.98 on Kvasir-SEG, and InceptionV3 and ResNet achieving approximately 0.99 AUC on CVC-ClinicDB [12,13]. Few-shot and transfer learning frameworks have been applied to address data scarcity [11]. However, most prior classification work focuses on binary tasks without coupling classification with segmentation or leveraging hierarchical subtype analysis.

### 2.3 Multi-Task and Hierarchical Learning

Multi-task learning—jointly optimizing related tasks through shared representations—has shown consistent benefits in medical imaging [15]. Hierarchical classification, which progressively refines predictions from coarse to fine-grained labels, has been explored in skin lesion analysis but remains underexplored in gastrointestinal pathology. PolypVision adopts a sequential hierarchical design where each stage's learned representations are transferred to the next, reflecting the clinical taxonomy of polyp assessment.

## 3. Datasets

PolypVision is trained and evaluated on three publicly available colorectal polyp datasets, selected to cover a range of clinical acquisition conditions, polyp types, and annotation modalities. Table 1 provides a summary of the datasets used.

**Table 1.** *Summary of datasets used for training and evaluation of PolypVision.*

| Dataset | Task(s) | Modality | Size | Classes / Details | Avail. | Year |
|---|---|---|---|---|---|---|
| PolypGen [15] | Segmentation, Detection | Image & Video | 3,762 images (1,537 single + 2,225 seq.) | Multi-center, 6 centers | Public | 2021 |
| Kvasir-SEG [16] | Segmentation, Detection | Image | 1,000 images | Pixel-wise masks | Public | 2019 |
| CVC-ClinicDB [17] | Segmentation | Image | 612 images (29 video sequences) | Frame-level annotations | Public | 2015 |
| ERCPMP [18] | Classification | Image & Video | 419 images, 37 videos | Endoscopic and pathologic annotations | Public | 2024 |

PolypGen [15] is a large-scale multi-center dataset comprising 3,762 positive polyp images (1,537 single-frame and 2,225 sequential frames) acquired across six clinical centers with varied patient populations, offering high diversity in imaging conditions and lesion appearance. Kvasir-SEG [16] provides 1,000 high-quality endoscopic polyp images with pixel-level segmentation masks, widely used as a benchmark for both detection and segmentation tasks. CVC-ClinicDB [17] contains 612 frames extracted from 29 colonoscopy video sequences, providing frame-level annotations suited for segmentation and classification evaluation.

For Stage 1 (binary classification), images are labeled as adenomatous or hyperplastic. For Stage 3 (adenoma subtype classification), adenomatous images are further labeled as tubular, tubulovillous, or villous. For Stage 2 (segmentation), pixel-wise polyp masks from Kvasir-SEG and CVC-ClinicDB are used. All images are resized to 224×224 pixels for classification stages and to variable resolution for segmentation. Stratified train/validation/test splits are applied to preserve class balance across all partitions.

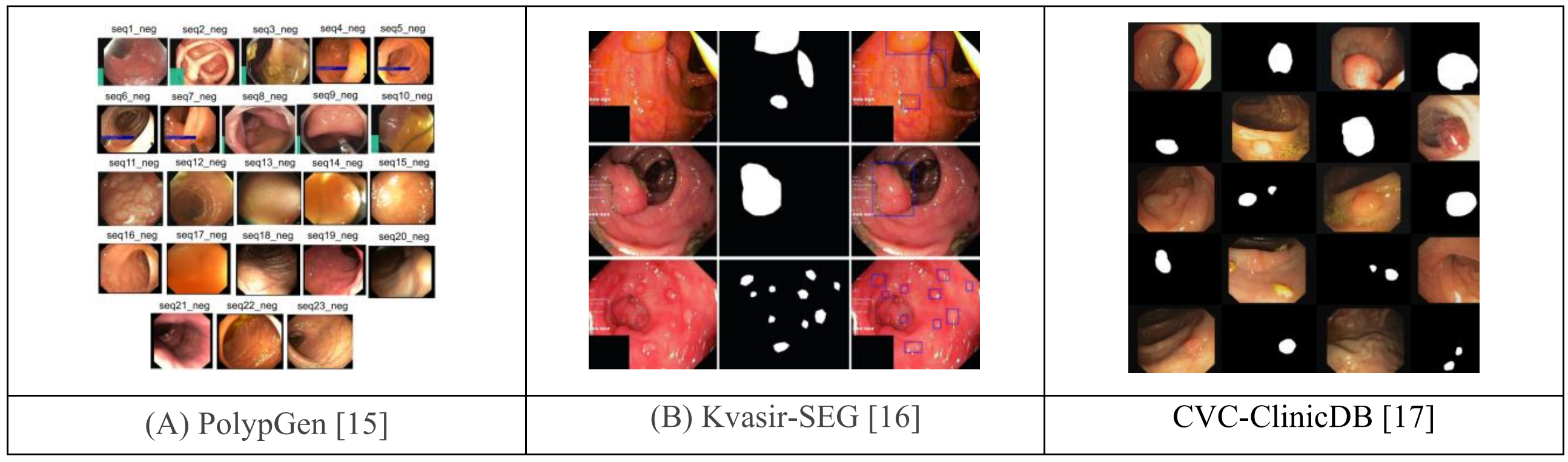

(A) PolypGen [15] | (B) Kvasir-SEG [16] | CVC-ClinicDB [17]

**Figure 1.** *Representative image samples from the three datasets used in PolypVision: (A) PolypGen — multi-center endoscopic frames; (B) Kvasir-SEG — polyp images with pixel-wise segmentation masks; (C) CVC-ClinicDB — colonoscopy video frames.*

# 4. Methods

## 4.1 System Architecture Overview

PolypVision implements a three-stage hierarchical pipeline in which each stage's trained backbone is transferred to the next stage, enabling progressive knowledge reuse. The system is deployed as a web application at https://polypvision.com (a DataBioX project), providing real-time inference with a free usage tier. Figure 2 illustrates the complete system architecture.

Hierarchical Polyp Classification & Segmentation Pipeline
STAGE 1: BINARY CLASSIFICATION
Input Image [B, 3, 224, 224]
EfficientNetV2-M Backbone 52.8M params
Global Avg Pool [B, feature_dim]
Dropout + Linear → [B, 2]
Output: Adenomatous or Hyperplastic Loss: Focal Loss
STAGE 2: SEGMENTATION (UNet++)
Input Image [B, 3, H, W]
ENCODER EfficientNetV2-M (From Stage 1) Frozen/Trainable
DECODER UNet++ Progressive Fusion Skip Connections
Output Mask [B, 1, H, W] Loss: Dice + BCE
STAGE 3: ADENOMA SUBTYPE
Input Image Adenomatous Only [B, 3, 224, 224]
EfficientNetV2-M From Stage 2 52.8M params
Dropout + Linear → [B, 3]
Output: Tubular Tubulovillous Villous Loss: Focal Loss
INFERENCE FLOW
Input Image
Stage 1 Classifier
Adenomatous
Adenomatous
Hyperplastic
Hyperplastic
Stage 3 Classifier (Subtype)
Villous
Tubular
Tubulovillous
Tubular
Tubulovillous
Villous
Legend
Backbone: EfficientNetV2-M (52.8M params)
Encoder: EfficientNetV2-M (transferred from previous stage)
Decoder: UNet++ with skip connections for segmentation
Transfer Learning: Stage1 → Stage2 Encoder, Stage2 Encoder → Stage3 Backbone
Losses: Stage1=FocalLoss, Stage2=Dice+BCE, Stage3=FocalLoss+MixUp

**Figure 2. PolypVision three-stage hierarchical pipeline.** *Stage 1: binary classification (Adenomatous vs. Hyperplastic) using EfficientNetV2-M with Focal Loss. Stage 2: polyp segmentation using UNet++ with the Stage 1 backbone as encoder (Dice + BCE Loss). Stage 3: adenoma subtype classification (Tubular / Tubulovillous / Villous) using EfficientNetV2-M transferred from Stage 2, trained with Focal Loss and MixUp augmentation. The inference flow (bottom right) shows the full decision pathway for an input image.*

### 4.2 Backbone: EfficientNetV2-M

EfficientNetV2-M [9] serves as the shared backbone across all three stages. With 52.8 million parameters, it combines compound scaling of depth, width, and resolution with Fused-MBConv blocks for efficient feature extraction. Pre-trained on ImageNet-21k, it provides rich hierarchical representations that transfer well to endoscopic imaging. For classification stages (1 and 3), a Global Average Pooling layer followed by Dropout and a linear classification head is appended. For Stage 2, the EfficientNetV2-M encoder is integrated into the UNet++ decoder.

### 4.3 Stage 1: Binary Classification

Stage 1 classifies input images [B, 3, 224, 224] as adenomatous or hyperplastic, and simultaneously predicts Paris classification (lesion morphology) and JNet classification (pit pattern and vascular pattern). The EfficientNetV2-M backbone extracts a global average-pooled feature vector [B, feature_dim], which passes through task-specific Dropout and linear heads to produce the three sets of outputs. Training uses Focal Loss to address class imbalance, with hyperparameters $\gamma = 2$ and $\alpha = 0.25$. The trained Stage 1 backbone is transferred as the encoder initialization for Stage 2.

### 4.4 Stage 2: Segmentation with UNet++

Stage 2 performs pixel-level polyp segmentation using UNet++ [10] with the Stage 1 EfficientNetV2-M backbone as encoder. The UNet++ decoder employs nested skip connections and progressive feature fusion, producing a binary segmentation mask [B, 1, H, W] at the original input resolution. In addition, Stage 2 predicts the recommended resection method (cold snare polypectomy, EMR, or ESD) based on lesion size, morphology, and segmented boundaries, using an auxiliary classification head. Training uses a combined Dice Loss and Binary Cross-Entropy (BCE) Loss for segmentation, with cross-entropy for resection prediction. The encoder can be frozen or fine-tuned depending on available data volume. After Stage 2 training, the encoder is transferred to Stage 3.

### 4.5 Stage 3: Adenoma Subtype Classification

Stage 3 receives only adenomatous images (filtered by Stage 1) and classifies them into three subtypes: tubular, tubulovillous, and villous. The EfficientNetV2-M backbone is initialized from the Stage 2 encoder, followed by Global Average Pooling, Dropout, and a linear head producing logits [B, 3]. Focal Loss is again used for class imbalance, and MixUp data augmentation is applied during training to improve generalization across visually similar adenoma subtypes.

### 4.6 Training Configuration

All stages use the AdamW optimizer with weight decay ($\lambda = 1\times10^{-4}$) and cosine learning rate scheduling with warm restarts. Classification stages apply layer-wise learning rate decay: higher rates ($1\times10^{-3}$) for newly initialized heads and lower rates ($1\times10^{-4}$ to $5\times10^{-5}$) for transferred backbone layers to prevent catastrophic forgetting. Early stopping based on validation AUC (classification) or Dice score (segmentation) selects optimal checkpoints. Standard augmentations include random flipping, rotation, and color jitter, with MixUp additionally applied at Stage 3.

## 5. Experiments and Results

### 5.1 Evaluation Protocol

Stage 1 and Stage 3 are evaluated using Area Under the ROC Curve (AUC), Accuracy, Precision, Recall, and F1-Score on held-out test sets. Stage 2 is evaluated using Dice Score, Intersection over Union (IoU), and Precision/Recall for segmentation quality. For comparison with prior detection methods, we additionally report mAP@50. All evaluations are performed on held-out test partitions not used during training or validation.

### 5.2 Detection and Segmentation Results

Table 2 compares PolypVision detection performance (mAP@50) against state-of-the-art methods on Kvasir-SEG and CVC-ClinicDB.

**Table 2.** *Detection performance (mAP@50) comparison on Kvasir-SEG and CVC-ClinicDB. PolypVision (Ours) rows are highlighted. "—" indicates result not available for that dataset.*

| Dataset | Model | mAP@50 |
|---|---|---|
| **Kvasir-SEG** | | |
| | YOLOv8-s | 91.16% |
| | CRH-YOLO | 90.7% |
| | YOLO-LAN | 96.19% |
| | YOLOv5 | 81.0% |
| | **PolypVision (Ours)** | **94.4%** |
| **CVC-ClinicDB** | | |
| | YOLOv8-m | 93.4% |
| | YOLOv4 (INT8) | 91.0% |
| | ResUNet++ (Detection) | 82.9% |
| | **PolypVision (Ours)** | — |

### 5.3 Classification Results

Table 3 compares PolypVision frame classification performance (AUC) against prior methods on Kvasir-SEG and CVC-ClinicDB.

**Table 3.** *Frame classification performance (AUC) comparison. PolypVision achieves AUC ~0.99 across both datasets, matching or exceeding state-of-the-art methods.*

| Dataset | Task | Model | AUC |
|---|---|---|---|
| **Kvasir-SEG** | | | |
| | Frame Classification | ResNet-50 / VGG16 | 0.91–0.98 |
| | Frame Classification | **PolypVision (Ours)** | **~0.99** |
| **CVC-ClinicDB** | | | |
| | Frame Classification | InceptionV3 / ResNet | ~0.99 |
| | Frame Classification | **PolypVision (Ours)** | **~0.99** |

### 5.4 Grad-CAM Interpretability Analysis

Gradient-weighted Class Activation Mapping (Grad-CAM) was applied to visualize the spatial regions driving Stage 1 and Stage 3 classification decisions. Grad-CAM computes the gradient of the class score with respect to the final convolutional layer feature maps, producing a heatmap that highlights discriminative image regions. In correctly classified cases, the model consistently focuses on the polyp lesion—surface texture, mucosal pit pattern, and lesion margins. Misclassified cases reveal attention dispersed over non-lesion areas, providing actionable feedback for targeted dataset augmentation and model refinement.

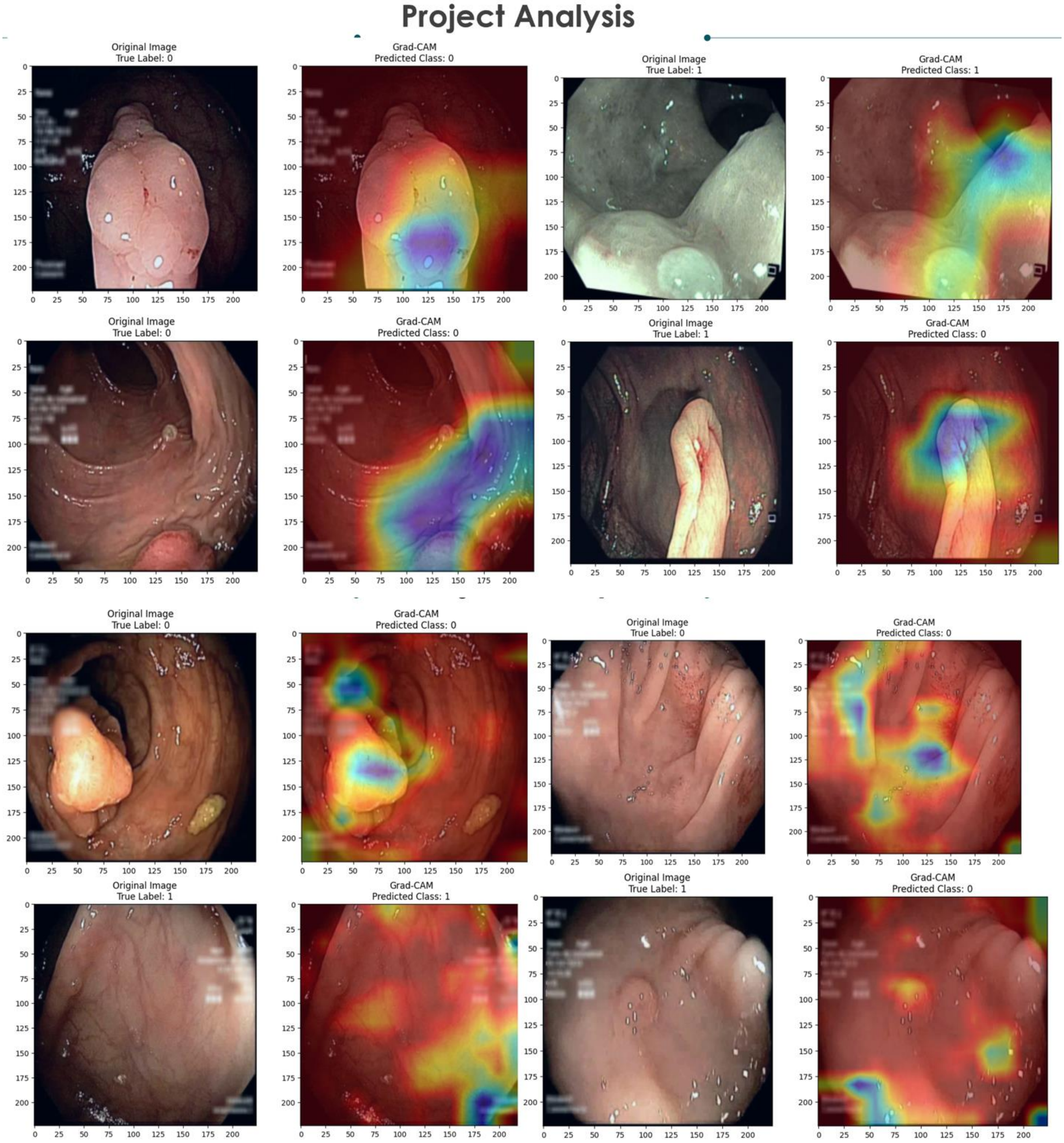

**Figure 3. Grad-CAM visualizations from the PolypVision Stage 1 classifier.** *Each panel pair shows the original endoscopic image (left) and the corresponding Grad-CAM heatmap (right). Warm colors (red/orange) indicate high model attention; cool colors (blue/green) indicate low activation. True Label 0 = hyperplastic; True Label 1 = adenomatous. The bottom-right panel illustrates a misclassified case (True Label 1, Predicted Class 0) in which model attention is dispersed over surrounding tissue rather than the polyp lesion.*

## 6. Discussion

PolypVision demonstrates that a hierarchical, transfer-learning-driven pipeline can achieve robust and comprehensive polyp analysis across multiple levels of clinical complexity—spanning morphological classification, endoscopic characterization, pixel-level segmentation, resection planning, and histological subtyping within a single unified framework.

The progressive transfer learning strategy—Stage 1 backbone → Stage 2 encoder → Stage 3 backbone—is a core strength of the framework. Shared representations reduce the effective data requirement per stage and encourage features simultaneously discriminative for classification and spatially coherent for segmentation. This mirrors the clinical workflow: identifying neoplastic lesions and characterizing morphology via Paris and JNet, then assessing boundaries and determining resection strategy, and finally determining histological subtype.

The simultaneous prediction of Paris and JNet classifications at Stage 1 substantially extends clinical utility beyond binary polyp typing. Paris characterizes lesion morphology (pedunculated, sessile, flat, depressed), while JNet provides pit pattern and vascular pattern assessment—both directly informing resection urgency and technique.

The addition of resection method recommendation at Stage 2 is a significant clinical contribution. Leveraging segmented lesion boundaries, size estimation, and Stage 1 morphological outputs, the framework recommends the appropriate resection technique—cold snare polypectomy, EMR, or ESD—reducing reliance on subjective judgment.

A critical practical advantage of PolypVision is its device-independent design. The framework operates on standard endoscopic images without requiring calibration to specific hardware, proprietary imaging modes, or narrow-band imaging systems, ensuring applicability across diverse endoscopy units.

Focal Loss addresses class imbalance in polyp datasets. MixUp augmentation at Stage 3 improves generalization across similar adenoma subtypes. Grad-CAM confirms that the model attends to clinically relevant features—mucosal pit patterns, surface texture, and lesion margins—essential for clinical acceptance.

A current limitation is that Stage 3 operates only on images filtered by Stage 1, so Stage 1 errors propagate downstream. Future work may explore end-to-end optimization or uncertainty-aware inference. Prospective validation on external datasets from diverse clinical centers is required before deployment.

## 7. Conclusion

We presented PolypVision, a three-stage hierarchical deep learning framework for comprehensive colorectal polyp analysis. PolypVision delivers a clinically complete assessment pipeline: Stage 1 simultaneously classifies polyps as adenomatous or hyperplastic and predicts Paris and JNet classifications; Stage 2 performs pixel-level segmentation and recommends the appropriate resection method; and Stage 3 further subclassifies adenomas into tubular, tubulovillous, and villous subtypes. The framework is device-independent, operating across diverse endoscopic imaging systems without hardware-specific adaptation. PolypVision achieves state-of-the-art performance—AUC ~0.99 for frame classification and mAP@50 of 94.4% on Kvasir-SEG—and is openly accessible at https://polypvision.com (a DataBioX initiative) with a free usage tier, with strong potential for integration into endoscopic practice following prospective clinical validation.

## Conflicts of Interest

The authors declare no conflicts of interest.

## Data Availability Statement

The PolypVision web application is freely accessible at https://polypvision.com (DataBioX, https://databiox.com), with a free usage tier for all users. All datasets used in this study are publicly available: ERCPMP [18] (https://doi.org/10.1186/s13104-024-07062-6), Kvasir-SEG (https://datasets.simula.no/kvasir-seg/), CVC-ClinicDB (http://polyp.grand-challenge.org/CVCClinicDB/), and PolypGen (https://www.synapse.org/#!Synapse:syn26376615).

## Author Biographies

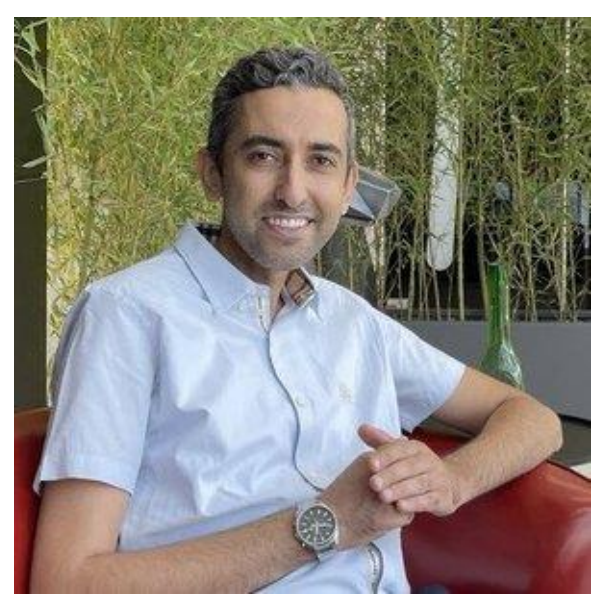

**Hamidreza Bolhasani, PhD,** is an AI/ML Researcher, Visiting Professor, and Founder and Chief Data Scientist at DataBioX (https://databiox.com). He holds a PhD in Computer Engineering with a focus on Artificial Intelligence and Computer Vision. His research spans Machine Learning, Deep Learning, Computer Vision, Biomedical AI, and Bioinformatics, with particular emphasis on AI applications in medical imaging and computational pathology. He led the overall research design and AI methodology for this work.

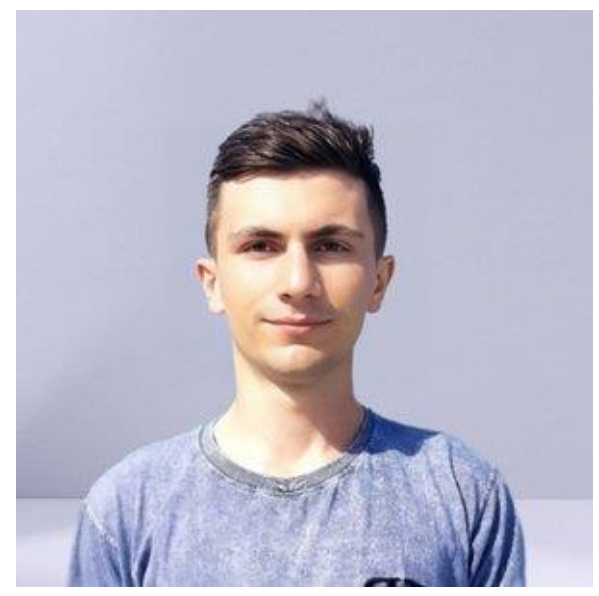

**Hamidreza Rastad,** is a final-year undergraduate student at Iran University of Science and Technology (IUST), Tehran, Iran. He specializes in artificial intelligence and deep learning, and has played a leading role in the design and development of deep learning models in this work and related projects in biomedical AI.

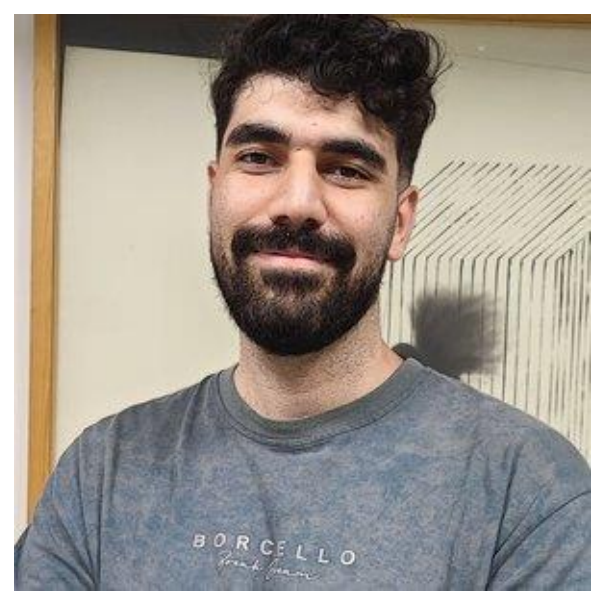

**Amir Mohammad Akbari,** is a student at Iran University of Science and Technology (IUST), Tehran, Iran. His contributions to this work span deep learning model development and UI/UX design, as well as support for AI system integration and implementation.

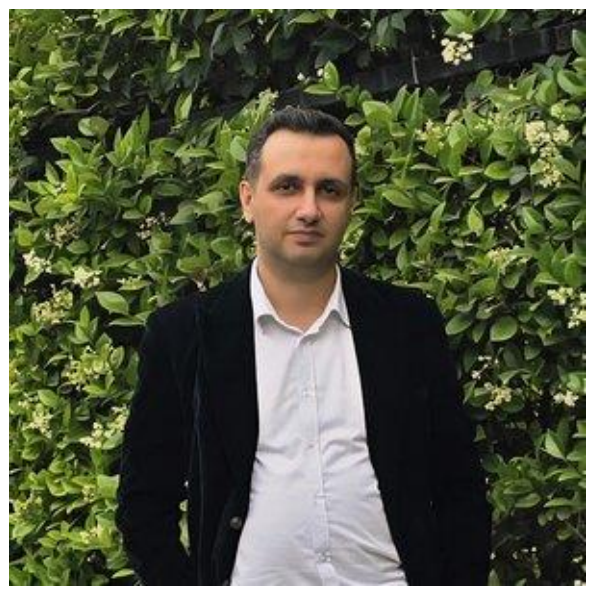

**Ata Khodami, MD,** is a specialist in Internal Medicine whose doctoral thesis addressed topics related to the subject of this work. His clinical interests include internal medicine and gastroenterology. His expertise contributed to the validation and clinical contextualization of the framework.

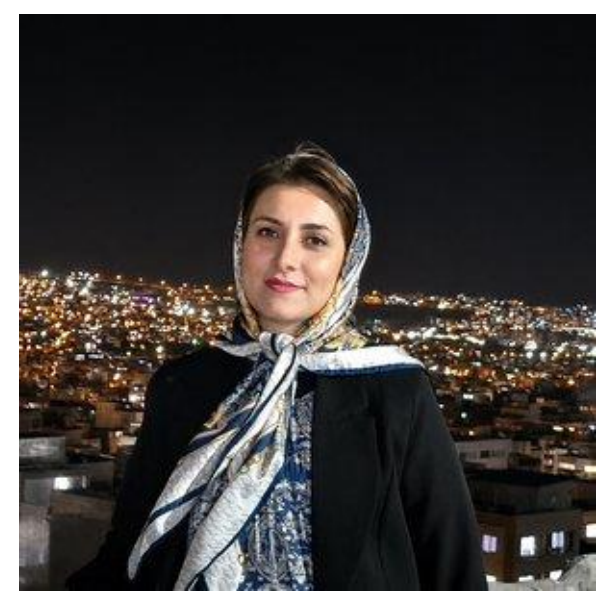

**Parnian Asadollahi, MD**, is a resident in Radiation Oncology at Shahid Beheshti University of Medical Sciences, Tehran, Iran. She contributed to this research through clinical data review, annotation support, and oncologic interpretation of gastrointestinal precancerous lesions.

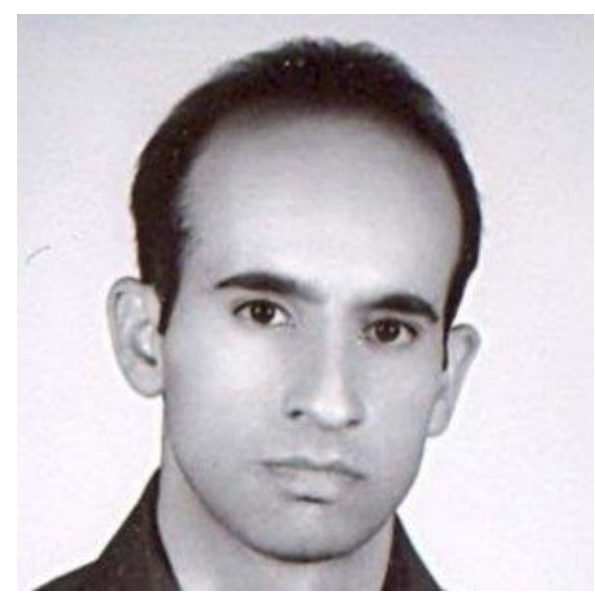

**Mohammad Tashakoripour,** is a Researcher at the Gastroenterology Department, Amiralam Hospital, Tehran University of Medical Sciences, Tehran, Iran. His work focuses on histopathological grading and validation of gastrointestinal precancerous lesions.

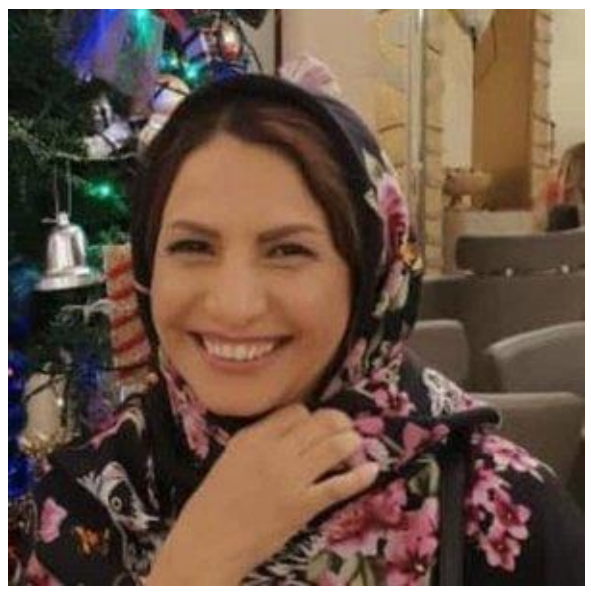

**Mojgan Forootan, MD,** is a Professor of Gastroenterology at the Gastroenterology and Liver Disease Research Center, Shahid Beheshti University of Medical Sciences, Tehran, Iran. She is interested in early detection and resection of GI precancerous lesions using AI. She is the corresponding author of this work.